\documentclass{article} 
\usepackage{iclr2017_conference,times,amsfonts}
\usepackage{hyperref}
\usepackage{url}
\usepackage{graphicx}
\usepackage{wrapfig}
\usepackage{amsmath}
\usepackage{booktabs}
\usepackage{latexsym}
\usepackage{multirow}
\usepackage{multicol}
\usepackage{amsmath}
\usepackage{subfigure}
\usepackage{float}

\makeatletter
\g@addto@macro\normalsize{%
  \abovedisplayskip 5pt plus 2pt minus 4pt%
  \belowdisplayskip \abovedisplayskip
  \abovedisplayshortskip 4pt plus3pt%
  \belowdisplayshortskip 4pt plus3pt minus3pt%
}
\makeatother

\title{Mode Regularized Generative Adversarial Networks}

\author{$^{\dagger}$Tong Che\thanks{Authors contributed equally.}, $^{\ddagger}$Yanran Li\footnotemark[1], $^{\dagger,\mathsection}$Athul Paul Jacob, $^{\dagger}$Yoshua Bengio, $^{\ddagger}$Wenjie Li\\
       $^{\dagger}$Montreal Institute for Learning Algorithms, Universit\'{e} de Montr\'{e}al, Montr\'{e}al, QC H3T 1J4, Canada\\
       $^{\ddagger}$Department of Computing, The Hong Kong Polytechnic University, Hong Kong\\
	  $^{\mathsection}$David R. Cheriton School of Computer Science, University Of Waterloo, Waterloo, ON N2L 3G1, Canada\\
       {\texttt \{tong.che,ap.jacob,yoshua.bengio\}@umontreal.ca} \\
       {\texttt \{csyli,cswjli\}@comp.polyu.edu.hk}
       }
\iclrfinalcopy 

\begin{document}

\maketitle

\begin{abstract}
Although Generative Adversarial Networks achieve state-of-the-art results on a variety of generative tasks, they are regarded as highly unstable and prone to miss modes. We argue that these bad behaviors of GANs are due to the very particular functional shape of the trained discriminators in high dimensional spaces, which can easily make training stuck or push probability mass in the wrong direction, towards that of higher concentration than that of the data generating distribution.

We introduce several ways of regularizing the objective, which can dramatically stabilize the training of GAN models. We also show that our regularizers can help the fair distribution of probability mass across the modes of the data generating distribution, during the early phases of training and thus providing a unified solution to the missing modes problem.
\end{abstract}

\section{Introduction}
Generative adversarial networks (GAN)~\citep{goodfellowgan} have demonstrated their potential on various tasks, such as image generation, image super-resolution, 3D object generation, and video prediction~\citep{DCGAN,photoresolution,sonderby2016amortised,nguyen2016ppgn,3dgan,ganvideo}. The objective is to train a parametrized function (the generator) which maps noise samples (e.g., uniform or Gaussian) to samples whose distribution is close to that of the data generating distribution. The basic scheme of the GAN training procedure is to train a discriminator which assigns higher probabilities to real data samples and lower probabilities to generated data samples, while simultaneously trying to move the generated samples towards the real data manifold using the gradient information provided by the discriminator. In a typical setting, the generator and the discriminator are represented by deep neural networks. 

Despite their success, GANs are generally considered as very hard to train due to training instability and sensitivity to hyper-parameters. On the other hand, a common failure pattern  observed while training GANs is the collapsing of large volumes of probability mass onto a few modes. Namely, although the generators produce meaningful samples, these samples are often from just a few modes (small regions of high probability under the data distribution). Behind this phenomenon is the missing modes problem, which is widely conceived as a major problem for training GANs: many modes of the data generating distribution are not at all represented in the generated samples, yielding a much lower entropy distribution, with less variety than the data generating distribution.

\begin{wrapfigure}{l}{0.45\textwidth}
\includegraphics[width=0.45\textwidth]{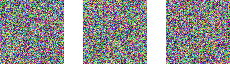} 
\caption{Samples with very high discrimination values (D=1.0) in DCGAN model trained on CelebA dataset.}
\label{fig:adv}
\end{wrapfigure}

This issue has been the subject of several recent papers proposing several tricks and new architectures to stabilize GAN's training and encourage its samples' diversity. However, we argue that a general cause behind these problems is the lack of control on the discriminator during GAN training. We would like to encourage the manifold of the samples produced by the generator to move towards that of real data, using the discriminator as a metric. However, even if we train the discriminator to distinguish between these two manifolds, we have no control over the shape of the discriminator function in between these manifolds. In fact, the shape of the discriminator function in the data space can be very non-linear with bad plateaus and wrong maxima and this can therefore hurt the training of GANs (Figure~\ref{fig:adv}). 

\newlength{\oldintextsep}
\setlength{\oldintextsep}{\intextsep}
\setlength\intextsep{0pt}
To remedy this problem, we propose a novel regularizer for the GAN training target. The basic idea is simple yet powerful: in addition to the gradient information provided by the discriminator, we want the generator to take advantage of other similarity metrics with much more predictable behavior, such as the $L_2$ norm. Differentiating these similarity metrics will provide us with more stable gradients to train our generator. Combining this idea with an approach meant to penalize the missing modes, we propose a family of additional regularizers for the GAN objective. We then design a set of metrics to evaluate the generated samples in terms of both the diversity of modes and the distribution fairness of the probability mass. These metrics are shown to be more robust in judging complex generative models, including those which are well-trained and collapsed ones.

Regularizers usually bring a trade-off between model variance and bias. Our results have shown that, when correctly applied, our regularizers can dramatically reduce model variance, stabilize the training, and fix the missing mode problem all at once, with positive or at the least no negative effects on the generated samples. We also discuss a variant of the regularized GAN algorithm, which can even improve sample quality as compared to the DCGAN baseline.

\section{Related Work}
The GAN approach was initially proposed by~\citet{goodfellowgan} where both the generator and the discriminator are defined by deep neural networks.

In~\citet{goodfellowgan}, the GAN is able to generate interesting local structure but globally incoherent images on various datasets. \citet{CGAN} enlarges GAN's representation capacity by introducing an extra vector to allow the generator to produce samples conditioned on other beneficial information. Motivated from this, several conditional variants of GAN has been applied to a wide range of tasks, including image prediction from a normal map~\cite{SSGAN}, image synthesis from text~\cite{reed2016generative} and edge map~\cite{pix2pix2016}, real-time image manipulation~\cite{zhu2016generative}, temporal image generation~\cite{zhou2016learning,saito2016temporal,vondrick2016generating}, texture synthesis, style transfer, and video stylization~\cite{li2016precomputed}. 

Researchers also aim at stretching GAN's limit to generate higher-resolution, photo-realistic images. \citet{LAPGAN} initially apply a Laplacian pyramid framework on GAN to generate images of high resolution. At each level of their LAPGAN, both the generator and the discriminator are convolutional networks. As an alternative to LAPGAN, \citet{DCGAN} successfully designs a class of deep convolutional generative adversarial networks which has led to significant improvements on unsupervised image representation learning. Another line of work aimed at improving GANs are through feature learning, including features from the latent space and image space. The motivation is that features from different spaces are complementary for generating perceptual and natural-looking images. With this perspective, some researchers use distances between learned features as losses for training objectives for generative models. \citet{VAEGAN} combine a variational autoencoder objective with a GAN and utilize the learned features from the discriminator in the GANs for better image similarity metrics. It is shown that the learned distance from the discriminator is of great help for the sample visual fidelity. Recent literature have also shown impressive results on image super-resolution to infer photo-realistic natural images for 4x upscaling factors~\cite{photoresolution,sonderby2016amortised,nguyen2016ppgn}. 

Despite these promising successes, GANs are notably hard to train. Although \citet{DCGAN} provide a class of empirical architectural choices that are critical to stabilize GAN's training, it would be even better to train GANs more robustly and systematically. \citet{improved} propose feature matching technique to stabilize GAN's training. The generator is required to match the statistics of intermediate features of the discriminator. Similar idea is adopted by~\citet{EBGAN}. In addition to feature distances, \citet{perceptual} found that the counterpart loss in image space further improves GAN's training stability. Furthermore, some researchers make use of information in both spaces in a unified learning procedure~\citep{ALI,AFL}. In \citet{ALI}, one trains not just a generator but also an encoder, and the discriminator is trained to distinguish between two joint distributions over image and latent spaces produced either by the application of the encoder on the training data or by the application of the generator (decoder) to the latent prior. This is in contrast with the regular GAN training, in which the discriminator only attempts to separate the distributions in the image space. Parallelly, \citet{metz2016unrolled} stabilize GANs by unrolling the optimization of discriminator, which can be considered as an orthogonal work with ours.

Our work is related to VAEGAN~\citep{VAEGAN} in terms of training an autoencoder or VAE jointly with the GAN model. However, the variational autoencoder (VAE) in VAEGAN is used to generate samples whereas our autoencoder based losses serves as a regularizer to penalize missing modes and thus improving GAN's training stability and sample qualities. We demonstrate detailed differences from various aspects in Appendix~\ref{appendix:vaegan}.

\section{Mode Regularizers for GANs}
The GAN training procedure can be viewed as a non-cooperative two player game, in which the discriminator $D$ tries to distinguish real and generated examples, while the generator $G$ tries to fool the discriminator by pushing the generated samples towards the direction of higher discrimination values. Training the discriminator $D$ can be viewed as training an evaluation metric on the sample space. Then the generator $G$ has to take advantage of the local gradient $\nabla \log D(G)$ provided by the discriminator to improve itself, namely to move towards the data manifold.

We now take a closer look at the root cause of the instabilities while training GANs. The discriminator is trained on both generated and real examples. As pointed out by \citet{goodfellowgan,LAPGAN,DCGAN}, when the data manifold and the generation manifold are disjoint (which is true in almost all practical situations), it is equivalent to training a characteristic function to be very close to 1 on the data manifold, and 0 on the generation manifold. In order to pass good gradient information to the generator, it is important that the trained discriminator produces stable and smooth gradients. However, since the discriminator objective does not directly depend on the behavior of the discriminator in other parts of the space, training can easily fail if the shape of the discriminator function is not as expected. As an example,\citet{LAPGAN} noted a common failure pattern for training GANs which is the vanishing gradient problem, in which the discriminator $D$ perfectly classifies real and fake examples, such that around the fake examples, $D$ is nearly zero. In such cases, the generator will receive no gradient to improve itself.\footnote{This problem exists even when we use $\log D(G(z))$ as target for the generator, as noted by \citet{LAPGAN} and our experiments.} 

Another important problem while training GANs is mode missing. In theory, if the generated data and the real data come from the same low dimensional manifold, the discriminator can help the generator distribute its probability mass, because the missing modes will not have near-0 probability under the generator and so the samples in these areas can be appropriately concentrated towards regions where $D$ is closer to 1. However, in practice since the two manifolds are disjoint, $D$ tends to be near 1 on all the real data samples, so large modes usually have a much higher chance of attracting the gradient of discriminator. For a typical GAN model, since all modes have similar $D$ values, there is no reason why the generator cannot collapse to just a few major modes. In other words,  since the discriminator's output is nearly 0 and 1 on fake and real data respectively, the generator is not penalized for missing modes. 

\subsection{Geometric Metrics Regularizer}
Compared with the objective for the GAN generator, the optimization targets for supervised learning are more stable from an optimization point of view. The difference is clear: the optimization target for the GAN generator is a learned discriminator. While in supervised models, the optimization targets are distance functions with nice geometric properties. The latter usually provides much easier training gradients than the former, especially at the early stages of training. 

Inspired by this observation, we propose to incorporate a supervised training signal as a regularizer on top of the discriminator target. Assume the generator $G(z): Z\rightarrow X$ generates samples by sampling first from a fixed prior distribution in space $Z$ followed by a deterministic trainable transformation $G$ into the sample space $X$. Together with $G$, we also jointly train an encoder $E(x):X\rightarrow Z$. Assume $d$ is some similarity metric in the data space, we add $\mathbb{E}_{x\sim p_d}[d(x,G\circ E(x))]$ as a regularizer, where $p_d$ is the data generating distribution. The encoder itself is trained by minimizing the same reconstruction error. 

In practice, there are many options for the distance measure $d$. For instance, the pixel-wise $L^2$ distance, or the distance of learned features by the discriminator~\citep{ALI} or by other networks, such as a VGG classifier. \citep{photoresolution}

The geometric intuition for this regularizer is straight-forward. We are trying to move the generated manifold to the real data manifold using gradient descent. In addition to the gradient provided by the discriminator, we can also try to match the two manifolds by other geometric distances, say, $L^s$ metric. The idea of adding an encoder is equivalent to first training a point to point mapping $G(E(x))$ between the two manifolds and then trying to minimize the expected distance between the points on these two manifolds. 

\subsection{Mode Regularizer}
In addition to the metric regularizer, we propose a mode regularizer to further penalize missing modes. In traditional GANs, the optimization target for the generator is the empirical sum $\sum_{i}\nabla_\theta \log D(G_\theta (z_i))$. The missing mode problem is caused by the conjunction of two facts: (1) the areas near missing modes are rarely visited by the generator, by definition, thus providing very few examples to improve the generator around those areas, and (2) both missing modes and non-missing modes tend to correspond to a high value of $D$, because the generator is not perfect so that the discriminator can take strong decisions locally and obtain a high value of $D$ even near non-missing modes.

\setlength{\oldintextsep}{\intextsep}
\setlength\intextsep{0pt}
\begin{wrapfigure}{l}{0.5\textwidth}
\includegraphics[width=0.5\textwidth]{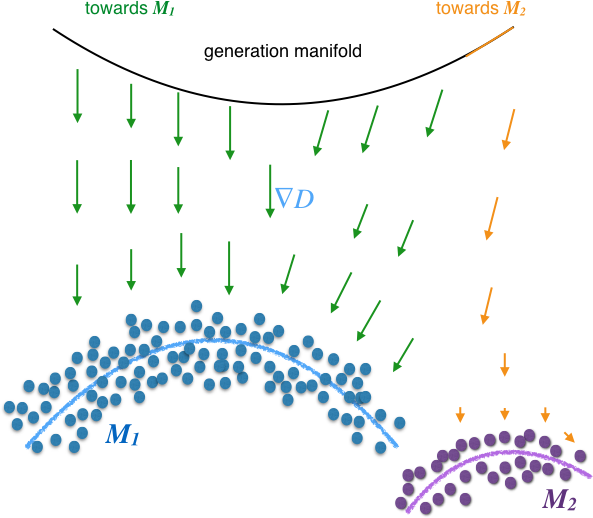} 
\caption{Illustration of missing modes problem.}
\label{fig:missingmodes}
\end{wrapfigure}

As an example, consider the situation in Figure ~\ref{fig:missingmodes}. For most $z$, the gradient of the generator $\nabla_\theta \log D(G_\theta(z))$ pushes the generator towards the major mode $M_1$. Only when $G(z)$ is very close to the mode $M_2$ can the generator get gradients to push itself towards the minor mode $M_2$. However, it is possible that such $z$ is of low or zero probability in the prior distribution $p_0$.

Given this observation, consider a regularized GAN model with the metric regularizer. Assume $M_0$ is a minor mode of the data generating distribution. For $x\in M_0$, we know that if $G\circ E$ is a good autoencoder, $G(E(x))$ will be located very close to mode $M_0$. Since there are sufficient training examples of mode $M_0$ in the training data, we add the mode regularizer $\mathbb{E}_{x\sim p_d}[\log D(G\circ E(x))]$ to our optimization target for the generator, to encourage $G(E(x))$ to move towards a nearby mode of the data generating distribution. In this way, we can achieve fair probability mass distribution across different modes. 

In short, our regularized optimization target for the generator and the encoder becomes:
\begin{eqnarray}
T_G = -\mathbb{E}_{z}[\log D(G(z))]+\mathbb{E}_{x\sim p_d}[\lambda_1 d(x,G\circ E(x)) + \lambda_2 \log D(G\circ E(x))] \\
T_E = \mathbb{E}_{x\sim p_d}[\lambda_1 d(x,G\circ E(x)) + \lambda_2 \log D(G\circ E(x))]
\end{eqnarray}

\subsection{Manifold-Diffusion Training for Regularized GANs}
\label{sec:MDGAN}
On some large scale datasets, CelebA for example, the regularizers we have discussed do improve the diversity of generated samples, but the quality of samples may not be as good without carefully tuning the hyperparameters. Here we propose a new algorithm for training metric-regularized GANs, which is very stable and much easier to tune for producing good samples. 

The proposed algorithm divides the training procedure of GANs into two steps: a manifold step and a diffusion step. In the manifold step, we try to match the generation manifold and the real data manifold with the help of an encoder and the geometric metric loss. In the diffusion step, we try to distribute the probability mass on the generation manifold fairly according to the real data distribution. 
%
%

An example of manifold-diffusion training of GAN (MDGAN for short) is as follows: we train a discriminator $D_1$ which separates between the samples $x$ and $G\circ E(x)$, for $x$ from the data, and we optimize $G$ with respect to the regularized GAN loss $\mathbb{E}[\log D_1(G\circ E(x))+\lambda d(x,G\circ E(x))]$ in order to match the two manifolds. In the diffusion step we train a discriminator $D_2$ between distributions $G(z)$ and $G\circ E(x)$, and we train $G$ to maximize $\log D_2(G(z))$. Since these two distributions are now nearly on the same low dimensional manifold, the discriminator $D_2$ provides much smoother and more stable gradients. The detailed training procedure is given in Appendix~\ref{appendix_algo}. See Figure~\ref{fig:celeba_samples} for the quality of generated samples. 
 
\subsection{Evaluation Metrics for Mode Missing}
In order to estimate both the missing modes and the sample qualities in our experiments, we used several different metrics for different experiments instead of human annotators. 

The inception score~\citep{improved} was considered as a good assessment for sample quality from a labelled dataset:
\begin{equation}
\exp \left(\mathbb{E}_{x} KL (p(y|\mathbf{x})||p^*(y))\right)
\end{equation}
Where $\mathbf{x}$ denotes one sample, $p(y|x)$ is the softmax output of a trained classifier of the labels, and $p^*(y)$ is the overall label distribution of generated samples. The intuition behind this score is that a strong classifier usually has a high confidence for good samples. However, the inception score is sometimes not a good metric for our purpose. Assume a generative model that collapse to a very bad image. Although the model is very bad, it can have a perfect inception score, because $p(y|x)$ can have a high entropy and $p^*(y)$ can have a low entropy. So instead, for labelled datasets, we propose another assessment for both visual quality and variety of samples, the MODE score:
\begin{equation}
\exp \left(\mathbb{E}_{x} KL (p(y|\mathbf{x})||p(y))-KL(p^*(y)||p(y))\right)
\end{equation}
 where $p(y)$ is the distribution of labels in the training data. According to our human evaluation experiences, the MODE score successfully measures two important aspects of generative models, i.e., variety and visual quality, in one metric.

However, in datasets without labels (LSUN) or where the labels are not sufficient to characterize every data mode (CelebA), the above metric does not work well. We instead train a third party discriminator between the real data and the generated data from the model. It is similar to the GAN discriminator {\em but is not used to train the generator}. We can view the output of the discriminator as an estimator for the quantity (See \citep{goodfellowgan} for proof):
\begin{equation}
D^*(s) \approx \frac{p_g(s)}{p_g(s)+p_d(s)}
\end{equation}
Where $p_g$ is the probability density of the generator and $p_d$ is the density of the data distribution. To prevent $D^*$ from learning a perfect 0-1 separation of $p_g$ and $p_d$, we inject a zero-mean Gaussian noise to the inputs when training $D^*$. After training, we test $D^*$ on the test set $T$ of the real dataset. If for any test sample $t\in T$, the discrimination value $D(t)$ is close to 1, we can conclude that the mode corresponding to $t$ is missing. In this way, although we cannot measure exactly the number of modes that are missing, we have a good estimator of the total probability mass of all the missing modes.

\section{Experiments}
\subsection{MNIST}
\setlength{\oldintextsep}{\intextsep}
\setlength\intextsep{0pt}
\begin{wraptable}{r}{6cm}
\caption{Grid Search for Hyperparameters.}
\label{table:grid}
\begin{tabular}{ll}\\\toprule
nLayerG & [2,3,4] \\
nLayerD & [2,3,4] \\
sizeG & [400,800,1600,3200] \\
sizeD & [256, 512, 1024] \\
dropoutD & [True,False] \\
optimG & [SGD,Adam] \\
optimD & [SGD,Adam] \\ 
lr &[1e-2,1e-3,1e-4] \\\bottomrule
\end{tabular}
\end{wraptable}
We perform two classes of experiments on MNIST. For the MNIST dataset, we can assume that the data generating distribution can be approximated with ten dominant modes, if we define the term ``mode'' here as a connected component of the data manifold.

\subsubsection{Grid Search for MNIST GAN Models}

In order to systemically explore the effect of our proposed regularizers on GAN models in terms of improving stability and sample quality, we use a large scale grid search of different GAN hyper-parameters on the MNIST dataset. The grid search is based on a pair of randomly selected loss weights: $\lambda_1=0.2$ and $\lambda_2=0.4$. We use the same hyper-parameter settings for both GAN and Regularized GAN, and list the search ranges in Table~\ref{table:grid}. Our grid search is similar to those proposed in~\citet{EBGAN}. Please refer to it for detailed explanations regarding these hyper-parameters. 

\begin{figure}[h]
\begin{center}
\includegraphics[width=0.8\textwidth]{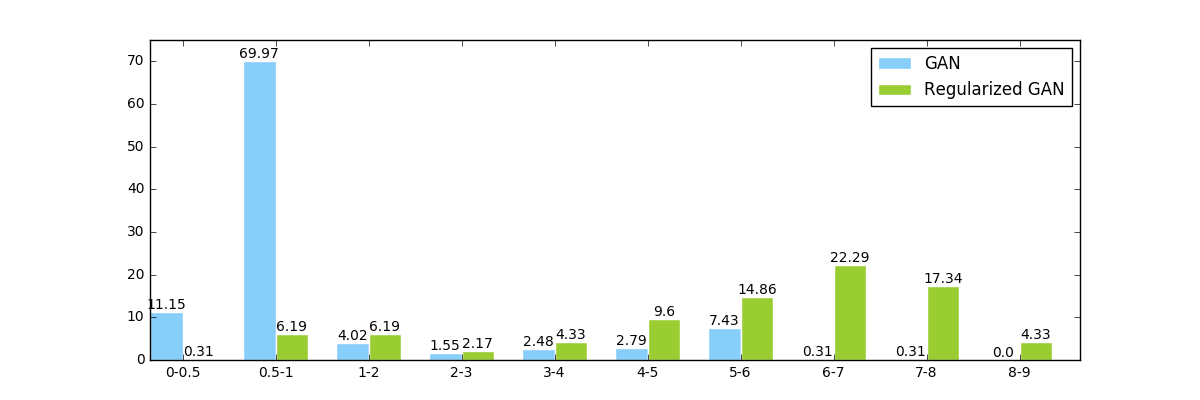}
\end{center}
\caption{The distributions of MODE scores for GAN and regularized GAN.}
\label{fig:modescore}
\end{figure}

For evaluation, we first train a 4-layer CNN classifier on the MNIST digits, and then apply it to compute the MODE scores for the generated samples from all these models. The resulting distribution of MODE score is shown in Figure~\ref{fig:modescore}. Clearly, our proposed regularizer significantly improves the MODE scores and thus demonstrates its benefits on stabilizing GANs and improving sample qualities.

\begin{figure}[h]
\begin{center}
\includegraphics[width=0.66\textwidth]{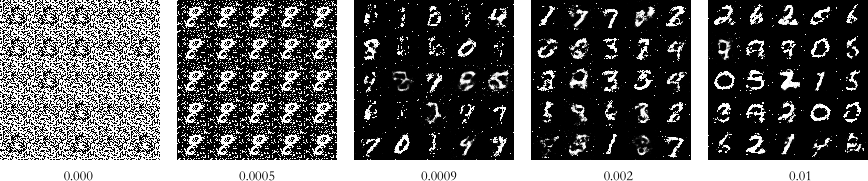}
\includegraphics[width=0.3\textwidth]{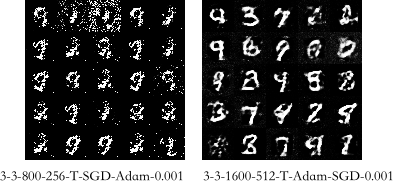}
\end{center}
\caption{(Left 1-5) Different hyperparameters for MNIST generation. The values of the $\lambda_1$ and $\lambda_2$ in our Regularized GAN are listed below the corresponding samples. (Right 6-7) Best samples through grid search for GAN and Regularized GAN.}
\label{fig:mnist}
\end{figure}

\vspace{0.5em}
To illustrate the effect of regularizers with different coefficients, we randomly pick an architecture and train it with different $\lambda_1=\lambda_2$. The results are shown in Figure~\ref{fig:mnist}.
\vspace{1em}

\subsubsection{Compositional MNIST data with 1000 modes}

In order to quantitatively study the effect of our regularizers on the missing modes, we concatenate three MNIST digits to a number in [0,999] in a single 64x64 image, and then train DCGAN as a baseline model on the 1000 modes dataset. The digits on the image are sampled with different probabilities, in order to test the model's capability to preserve small modes in generation. We again use a pre-trained classifier for MNIST instead of a human to evaluate the models. 

\vspace{1em}
\begin{table}[h]
\begin{center}
\caption{Results for Compositional MNIST with 1000 modes. The proposed regularization (Reg-DCGAN) allows to substantially reduce the number of missed modes as well as the KL divergence that measures the plausibility of the generated samples (like in the Inception score).}
\label{table:mnist_seq}
\begin{tabular}{r |cc cc cc cc}\toprule
\multirow{2}{*}{} & \multicolumn{2}{c}{Set 1} & \multicolumn{2}{c}{Set 2} & \multicolumn{2}{c}{Set 3} & \multicolumn{2}{c}{Set 4} \\
& {\#Miss} & KL & {\#Miss} & KL & {\#Miss} & KL & {\#Miss} & KL \\\midrule
DCGAN & 204.7 & 77.9 & 204.3 & 60.2 & 103.4 & 75.9 & 89.3 &77.8\\ \midrule
\bf{Reg-DCGAN} & \bf{32.1} & \bf{62.3} & \bf{71.5} & \bf{58.9} & \bf{42.7} & \bf{68.4} & \bf{31.6} & \bf{67.8}\\
\bottomrule
\end{tabular}
\end{center}
\end{table}

The performances on the compositional experiment are measured by two metrics. \#Miss represents the classifier-reported number of missing modes, which is the size of the set of numbers that the model never generates. KL stands for the KL divergence between the classifier-reported distribution of generated numbers and the distribution of numbers in the training data (as for the Inception score). The results are shown in Table~\ref{table:mnist_seq}. With the help of our proposed regularizer, both the number of missing modes and KL divergence drop dramatically among all the sets of the compositional MNIST dataset, which again proves the effectiveness of our regularizer for preventing the missing modes problem. 

\subsection{CelebA}
To test the effectiveness of our proposal on harder problems, we implement an encoder for the DCGAN algorithm and train our model with different hyper-parameters together with the DCGAN baseline on the CelebA dataset. We provide the detailed architecture of our regularized DCGAN in Appendix~\ref{appendix_arch}.

\subsubsection{Missing Modes Estimation on CelebA}
\label{sec:missing_modes}
We also employ a third party discriminator trained with injected noise as a metric for missing mode estimation. To implement this, we add noise in the input layer in the discriminator network. For each GAN model to be estimated, we independently train this noisy discriminator, as mode estimator, with the same architecture and hyper-parameters on the generated data and the training data. We then apply the mode estimator to the test data. The images which have high mode estimator outputs can be viewed as on the missing modes. 

\vspace{1em}
\begin{table}[h]
\begin{center}
\caption{Number of images on the missing modes on CelebA estimated by a third-party discriminator. The numbers in the brackets indicate the dimension of prior $z$. $\sigma$ denotes the standard deviation of the added Gaussian noise applied at the input of the discriminator to regularize it. MDGAN achieves a very high reduction in the number of missing modes, in comparison to other methods
.}
\label{table:celeba_mode}
\begin{tabular}{c| cc c c c}\toprule
 $\sigma$& DCGAN (100) & DCGAN (200) & Reg-GAN (100) & Reg-GAN (200) & MDGAN (200) \\\midrule
3.5 & 5463  & 17089 & 754 & 3644  & \bf{74}\\\midrule
4.0 & 590 &15832 & 42 &391 &\bf{13} \\
\bottomrule
\end{tabular}
\end{center}
\end{table}
\vspace{1em}

The comparison result is shown in Table~\ref{table:celeba_mode}. Both our proposed Regularized-GAN and MDGAN outperform baseline DCGAN models on all settings. Especially, MDGAN suppresses other models, showing its superiority on modes preserving. We also find that, although sharing the same architecture, the DCGAN with 200-dimensional noise performs quite worse than that with 100-dimensional noise as input. On the contrary, our regularized GAN performs more consistently.

\begin{figure}[h]
\includegraphics[width=\textwidth]{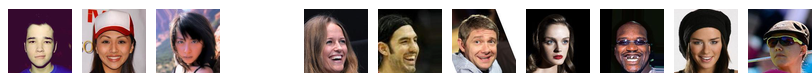}
\caption{Test set images that are on missing mode. Left: Both MDGAN and DCGAN missing. Right: Only DCGAN missing.}
\label{fig:miss_samples}
\end{figure}

To get a better understanding of the models' performance, we want to figure out when and where these models miss the modes. Visualizing the test images associated with missed modes is instructive. In Figure~\ref{fig:miss_samples}, the left three images are missed by all models. It is rare to see in the training data the cap in the second image and the type of background in the third, which thus can be viewed as small modes under this situation. These three images should be considered as the hardest test data for GAN to learn. Nonetheless, our best model, MDGAN still capture certain small modes. The seven images on the right in Figure~\ref{fig:miss_samples} are only missed by DCGAN. The sideface, paleface, black, and the berets are special attributes among these images, but our proposed MDGAN performs well on all of them.

\vspace{1em}
\vspace{1em}

\subsubsection{Qualitative Evaluation of Generated Samples}
\label{sec:sample_quality}

\begin{figure}[h]
\includegraphics[width=\textwidth]{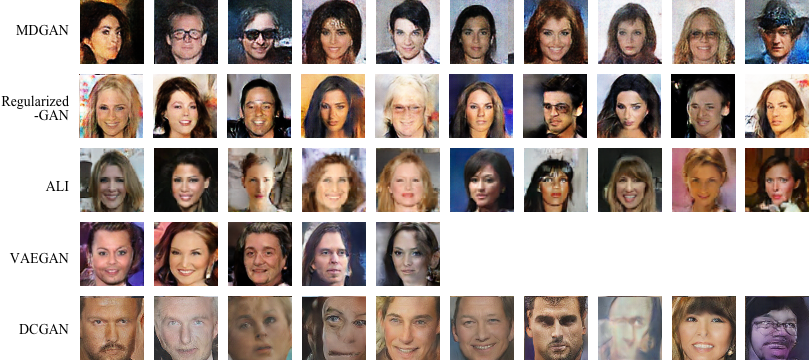}
\caption{Samples generated from different generative models. For each compared model, we directly take ten decent samples reported in their corresponding papers and code repositories. Note how MDGAN samples are both globally more coherent and locally have sharp textures.}
\label{fig:celeba_samples}
\end{figure}

After quantitative evaluation, we manually examine the generated samples by our regularized GAN to see whether the proposed regularizer has side-effects on sample quality. We compare our model with ALI~\citep{ALI}, VAEGAN~\citep{VAEGAN}, and DCGAN~\citep{DCGAN} in terms of sample visual quality and mode diversity. Samples generated from these models are shown in Figure~\ref{fig:celeba_samples}\footnote{For fair comparison, we also recommend readers to refer to the original papers~\cite{ALI,VAEGAN,DCGAN} for the reported samples of the compared. The ALI samples are from \url{https://github.com/IshmaelBelghazi/ALI/blob/master/paper/celeba_samples.png} and we reverted them to the original 64x64 size. The DCGAN samples are from \url{https://github.com/Newmu/dcgan_code/}}.

\vspace{1em}
\vspace{1em}

Both MDGAN and Regularized-GAN generate clear and natural-looking face images. Although ALI's samples are plausible, they are sightly deformed in comparison with those from MDGAN. The samples from VAEGAN and DCGAN seem globally less coherent and locally less sharp. 

As to sample quality, it is worth noting that the samples from MDGAN enjoy fewer distortions. With all four other models, the majority of generated samples suffer from some sort of distortion. However, for the samples generated by MDGAN, the level of distortion is lower compared with the other four compared models. We attribute it to the help of the autoencoder as the regularizer to alter the generation manifolds. In this way, the generator is able to learn fine-grained details such as face edges. As a result, MDGAN is able to reduce distortions. 

\vspace{1em}
\begin{figure}[h]
\includegraphics[width=\textwidth]{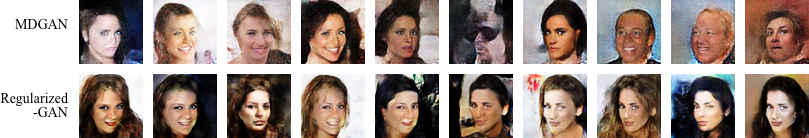}
\caption{Sideface samples generated by Regularized-GAN and MDGAN.}
\label{fig:sideface}
\end{figure}

In terms of missing modes problem, we instructed five individuals to conduct human evaluation on the generated samples. They achieve consensus that MDGAN wins in terms of mode diversities. Two people pointed out that MDGAN generates a larger amount of samples with side faces than other models. We select several of these side face samples in Figure~\ref{fig:sideface}. Clearly, our samples maintain acceptable visual fidelity meanwhile share diverse modes. Combined with the above quantitative results, it is convincing that our regularizers bring benefits for both training stability and mode variety without the loss of sample quality. 

\section{Conclusions}
Although GANs achieve state-of-the-art results on a large variety of unsupervised learning tasks, training them is considered highly unstable, very difficult and sensitive to hyper-parameters, all the while, missing modes from the data distribution or even collapsing large amounts of probability mass on some modes. Successful GAN training usually requires large amounts of human and computing efforts to fine tune the hyper-parameters, in order to stabilize training and avoid collapsing. Researchers usually rely on their own experience and published tricks and hyper-parameters instead of systematic methods for training GANs. 

We provide systematic ways to measure and avoid the missing modes problem and stabilize training with the proposed autoencoder-based regularizers. The key idea is that some geometric metrics can provide more stable gradients than trained discriminators, and when combined with the encoder, they can be used as regularizers for training. These regularizers can also penalize missing modes and encourage a fair distribution of probability mass on the generation manifold. 

\section*{Acknowledgements}
We thank Naiyan Wang, Jianbo Ye, Yuchen Ding, Saboya Yang for their GPU support. We also want to thank Huiling Zhen for helpful discussions, Junbo Zhao for providing the details of grid search experiments on the EBGAN model, as well as Anders Boesen Lindbo Larsen for kindly helping us on running VAEGAN experiments. We appreciate for the valuable suggestions and comments from the anonymous reviewers. The work described in this paper was partially supported by NSERC, Calcul Quebec, Compute Canada, the Canada Research Chairs, CIFAR, National Natural Science Foundation of China (61672445 and 61272291), Research Grants Council of Hong Kong (PolyU 152094/14E), and The Hong Kong Polytechnic University (G-YBP6).

\bibliography{iclr2017_conference}
\bibliographystyle{iclr2017_conference}

\newpage
\appendix
\section{Appendix: Pseudo code for MDGAN} 
\label{appendix_algo}
In this Appendix, we give the detailed training procedure of an MDGAN example we discuss in Section~\ref{sec:MDGAN}.
\vspace{0.5cm}
\begin{figure}[h]
\centering
\begin{tabular}{p{\textwidth}}\hline\hline
\vspace{0.04cm}
\bf{Manifold Step:}\\
    1. Sample $\{\mathbf{x}_1,\mathbf{x}_2,\cdots \mathbf{x}_m\}$ from data generating distribution $p_{data}(x)$.\\
    2. Update discriminator $D_1$ using SGD with gradient ascent: 
    \begin{equation*}
    \nabla_{\theta_d^1} \frac{1}{m}\sum_{i=1}^m [\log D_1(\mathbf{x}_i)+\log(1-D_1(G(E(\mathbf{x}_i))))]
    \end{equation*}\\
    3. Update generator $G$ using SGD with gradient ascent:
    \begin{equation*}
    \nabla_{\theta_g} \frac{1}{m}\sum_{i=1}^m [\lambda \log D_1(G(E(\mathbf{x}_i)))-||\mathbf{x}_i-G(E(\mathbf{x}_i))||^2]
    \end{equation*}
\bf{Diffusion Step:}\\
    4. Sample $\{\mathbf{x}_1,\mathbf{x}_2,\cdots \mathbf{x}_m\}$ from data generating distribution $p_{data}(x)$.\\
    5. Sample $\{\mathbf{z}_1,\mathbf{z}_2,\cdots \mathbf{z}_m\}$ from prior distribution $p_{\sigma}(z)$. \\
    6. Update discriminator $D_2$ using SGD with gradient ascent: 
    \begin{equation*}
    \nabla_{\theta_d^2} \frac{1}{m}\sum_{i=1}^m [\log D_2(G(E(\mathbf{x}_i)))+\log(1-D_2(\mathbf{z}_i))]
    \end{equation*}\\
    7. Update generator $G$ using SGD with gradient ascent:
    \begin{equation*}
    \nabla_{\theta_g} \frac{1}{m}\sum_{i=1}^m [\log D_2(G(\mathbf{z}_i))]
    \end{equation*}\\
\hline\hline
\end{tabular}
\caption{The detailed training procedure of an MDGAN example.}
\end{figure}

\section{Appendix: Architecture For Experiments}
\label{appendix_arch}
We use similar architectures for Compositional MNIST and CelebA experiments. The architecture is based on that found in DCGAN~\cite{DCGAN}. Apart from the discriminator and generator which are the same as DCGAN, we add an encoder which is the "inverse" of the generator, by reversing the order of layers and replacing the de-convolutional layers with convolutional layers. 

One has to pay particular attention to batch normalization layers. In DCGAN, there are batch normalization layers both in the generator and the discriminator. However, two classes of data go through the batch normalization layers in the generator. One come from sampled noise $z$, the other one come from the encoder. In our implementation, we separate the batch statistics for these two classes of data in the generator, while keeping the parameters of BN layer to be shared. In this way, the batch statistics of these two kinds of batches cannot interfere with each other. 

\section{Appendix: Additional synthesized experiments}
To demonstrate the effectiveness of mode-regularized GANs proposed in this paper, we train a very simple GAN architecture on synthesized 2D dataset, following~\citet{metz2016unrolled}.

The data is sampled from a mixture of 6 Gaussians, with standard derivation of 0.1. The means of the Gaussians are placed around a circle with radius 5. The generator network has two ReLU hidden layers with 128 neurons. It generates 2D output samples from 3D uniform noise from [0,1]. The discriminator consists of only one fully connected layer of ReLU neurons, mapping the 2D input to a real 1D number. Both networks are optimized with the Adam optimizer with the learning rate of 1e-4.

\begin{figure}[H]
\includegraphics[width=\textwidth]{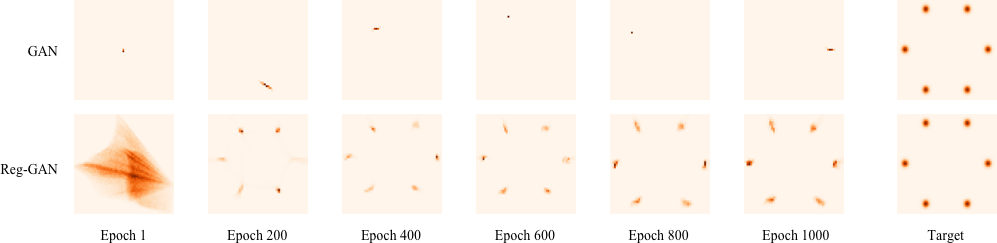}
\caption{Comparison results on a toy 2D mixture of Gaussians
dataset. The columns on the left shows heatmaps of the generator distributions as the number of training epochs increases, whereas the rightmost column presents the target, the original data distribution. The
top row shows standard GAN result. The generator has a hard time oscillating among the modes of the data
distribution, and is only able to ``recover'' a
single data mode at once. In contrast, the bottom row shows results of our regularized GAN. Its generator quickly captures the underlying multiple modes and fits the target distribution. }
\label{fig:synthetic}
\end{figure}

In the regularized version, we choose $\lambda_1=\lambda_2 = 0.005$. The comparison between the generator distribution from standard GAN and our proposed regularized GAN are shown in Figure~\ref{fig:synthetic}.

\vspace{1em}
\vspace{1em}

\section{Appendix: Comparison with VAEGAN}
\label{appendix:vaegan}
In this appendix section, we demonstrate the effectiveness and uniqueness of mode-regularized GANs proposed in this paper as compared to~\citet{VAEGAN} in terms of its theoretical difference, sample quality and number of missing modes. 

With regard to the theoretical difference, the optimization of VAEGAN relies on the probabilistic variational bound, namely $p(x)\geq \mathbb{E}_{q(z|x)}[\log p(x|z)]-\text{KL}(q(z|x)||p(z))$. This variational bound together with a GAN loss is optimized with several assumptions imposed in VAEGAN:
\begin{enumerate}
\item In general, VAE is based on the assumption that the true posterior $p(z|x)$ can be well approximated by factorized Gaussian distribution $q$.
\item As to VAEGAN, It is also assumed that the maximum likelihood objectives does not conflict with GAN objective in terms of probabilistic framework.
\end{enumerate}

The first assumption does not necessarily hold for GANs. We have found that in some trained models of DCGANs, the real posterior $p(z|x)$ is even not guaranteed to have only one mode, not to mention it is anything close to factorized Gaussian. We believe that this difference in probabilistic framework is an essential obstacle when one tries to use the objective of VAEGAN as a regularizer. However, in our algorithm, where we use a plain auto-encoder instead of VAE as the objective. Plain auto-encooders works better than VAE for our purposes because as long as the model $G(z)$ is able to generate training samples, there always exists a function $E^*(x)$ such that $G(E(x))= x$. Our encoder can therefore be viewed as being trained to approximate this real encoder $E^*$. There are no conflicts between a good GAN generator and our regularization objective. Hence, our objectives can be used as regularizers for encoding the prior knowledge that good models should be able to generate the training samples. This is why our work is essentially different from VAEGAN. In our experiments, we also believe that this is the reason why VAEGAN generates worse samples than a carefully tuned regularized GANs.

In terms of sample quality and missing modes, we run the official code of VAEGAN~\footnote{\url{https://github.com/andersbll/autoencoding_beyond_pixels}} with their default setting. We train VAEGAN for 30 epochs~\footnote{Note that we also trained 20-epoch version of VAEGAN, however the samples seemed worse.} and our models for only 20 epochs. For fairness, their model was run 3 times and the trained model with the best sample visual quality was taken for the comparison. 

\begin{figure}[H]
\includegraphics[width=\textwidth]{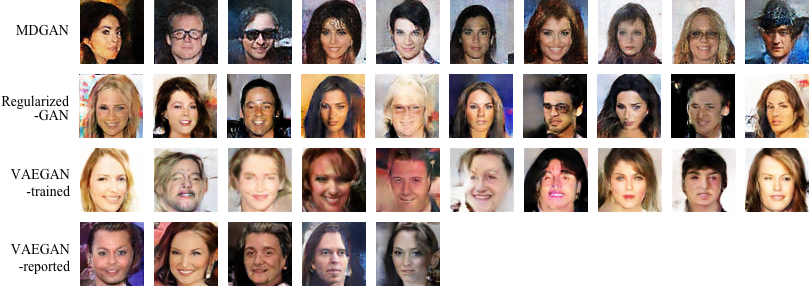}
\label{fig:vaegan_compare}
\caption{Samples generated by our models and VAEGAN. The third line are samples generated by our self-trained VAEGAN model, with default settings. The last line are generated samples reported in the original VAEGAN paper. We depict both of them here for a fair comparison.}
\label{fig:vaegan}
\end{figure}

The generated samples are shown in Figure~\ref{fig:vaegan}. The most obvious difference between our samples and VAEGAN's samples is the face distortion, which is consistent with our experimental results in Section~\ref{sec:sample_quality}. We conjecture that the distortions of VAEGAN's samples are due to the conflicts between the two objectives, as we present above. In other words, the way we introduce auto-encoders as regularizers for GAN models is different from VAEGAN's. The difference is that the second assumption mentioned above is not required in our approaches. In our framework, the auto-encoders helps alter the generation manifolds, leading to fewer distortions in fine-grained details in our generated samples.

\vspace{1em}
\vspace{1em}

In terms of the missing modes problem, we use the same method described in Section~\ref{sec:missing_modes} for computing the number of images with missing modes. The results are shown below.

\vspace{1em}
\begin{table}[h]
\begin{center}
\caption{Number of images on the missing modes on CelebA estimated by a third-party discriminator. The numbers in the brackets indicate the dimension of prior $z$. $\sigma$ denotes the standard deviation of the added Gaussian noise applied at the input of the discriminator to regularize it. MDGAN achieves a very high reduction in the number of missing modes, in comparison to VAEGAN.}
\label{table:vaegan_mode}
\begin{tabular}{c| c c c c}\toprule
 $\sigma$& VAEGAN (100) & Reg-GAN (100) & Reg-GAN (200) & MDGAN (200) \\\midrule
3.5 & 9720 & 754 & 3644  & \bf{74}\\\midrule
4.0 & 5862 & 42 &391 &\bf{13} \\
\bottomrule
\end{tabular}
\end{center}
\end{table}
\vspace{1em}
We see that using our proposed regularizers results in a huge drop in the number of missing modes. We conjecture that the reason why VAEGAN performs very bad in our metric for missing modes is because the samples generated are of low quality, so the discriminator classifies the samples as ``not on mode''. Namely, the data generated is too far away from many real data modes. Essentially if a model generates very bad samples, we can say that the model misses all or most modes.

To conduct more fair evaluation between VAEGAN and our methods, we also perform a blind human evaluation. Again we instructed five individuals to conduct this evaluation of sample variability. Without telling them which is generated by VAEGAN and which is generated by our methods, four people agree that our method wins in terms of sample diversity. One person thinks the samples are equally diverse. 

In conclusion, we demonstrate that our proposed mode-regularized GANs, i.e., Reg-GAN and MDGAN, are different from VAEGAN theoretically as discussed above. Such differences empirically result in better sample quality and mode preserving ability, which are our main contributions.
\end{document}